# Understanding Public Perception of Crime in Bangladesh: A Transformer-Based Approach with Explainability

Fatema Binte Hassan[1], Md Al Jubair[2], Mohammad Mehadi Hasan[3],
Tahmid Hossain[4], S M Mehebubur Rahman Khan Shuvo[5], Mohammad Shamsul Arefin[6]
[1,2,6]Department of CSE, Chittagong University of Engineering and Technology, Chittagong, Bangladesh
[3]Department of CSE, Bangladesh University of Business and Technology, Dhaka, Bangladesh
[4,5]Department of CSE, European University of Bangladesh, Dhaka, Bangladesh
Email: [1]fatemabintehassan@gmail.com, [2]aljubairpollob@gmail.com, [3]hasanbdmehadi@gmail.com,
[4]htahmid995@gmail.com, [5]shuvookhan24@gmail.com, [6]sarefin@cuet.ac.bd

*Abstract*—In recent years, social media platforms have become prominent spaces for individuals to express their opinions on ongoing events, including criminal incidents. As a result, public sentiment can shift dynamically over time. This study investigates the evolving public perception of crime-related news by classifying user-generated comments into three categories: positive, negative, and neutral. A newly curated dataset comprising 28,528 Bangla-language social media comments was developed for this purpose. We propose a transformer-based model utilizing the XLM-RoBERTa Base architecture, which achieves a classification accuracy of 97%, outperforming existing state-of-the-art methods in Bangla sentiment analysis. To enhance model interpretability, explainable AI technique is employed to identify the most influential features driving sentiment classification. The results underscore the effectiveness of transformer-based models in processing low-resource languages such as Bengali and demonstrate their potential to extract actionable insights that can support public policy formulation and crime prevention strategies.

*Index Terms*—XLM-RoBERTa; NLP; Sentiment Analysis; Bangla Sentiment Analysis; XAI

## I. Introduction

As social media sites change quickly, people are adjusting how they communicate and share information. In Bangladesh, platforms like Facebook, Twitter, and YouTube have become key places for people to share their thoughts on many issues, including crime. Given the increasing prevalence of crime in the region, public sentiment analysis through social media has gained considerable interest as a means of understanding social perceptions of crime and law enforcement agencies [1].

It has become an essential technique for systematically analyzing public opinion across large-scale digital platforms, particularly in the context of social media discourse. Most existing sentiment analysis studies primarily focus on English-language text. On the other hand, due to the scarcity of Bangla text, its complex structure, and the extensive use of informal expression, Bengali sentiment analysis work is limited [2]. In Bangladesh, most social media content is written in Bangla and often includes slang and mixed languages. This makes it difficult to analyze people's feelings about the content. These factors emphasize the suitable methodologies for sentiment analysis of Bangla text [3].

Crime, which is an illegal act that can lead to punishment, remains a big issue in Bangladesh. The Bangladesh Tribunal reports that the country successfully prosecutes many hacking cases, with 925 cases in 2018 [4] and 130 more in the first two months of 2019 (Dhaka Tribune, 2019) [5]. Crimes like kidnapping, murder, blackmail, and rape are common and often happen because of power conflicts in local, political, or social situations. Political abuse of power makes these problems worse, putting people, especially women and children, at greater risk.

Bangladesh has also struggled with terrorist attacks carried out by local entities, with the government making significant efforts in recent years to neutralize the threat (Terrorism in Bangladesh, 2020) [6]. This study introduces a transformer-based model designed to tackle two key challenges in natural language processing. First, it aims to analyze public discourse concerning the increasing crime rates in society. Second, it addresses the linguistic complexities inherent in processing the Bengali language. The proposed model effectively classifies sentiment—categorized as positive, negative, and neutral—in user-generated social media comments concerning crime-related events. The aim of this study is to offer beneficial suggestions to law enforcement agencies, policymakers, and researchers, demonstrating the potential of pretraining models for sentiment analysis in underrepresented languages such as Bengali.

The primary contributions are outlined as follows:

- A novel dataset comprising user-generated comments on crime-related news articles in the Bengali language has been developed to facilitate sentiment analysis within a low-resource linguistic context.

- Applied translation-based augmentation to enrich the dataset and reduce sparsity.
- Employed the XLM-RoBERTa Base transformer for robust sentiment classification.
- The proposed model outperforms leading Bengali sentiment analysis methods, effectively handling linguistic and contextual nuances.
- To enhance transparency, used explainable AI [7] techniques to interpret and validate model decisions.

Similar efforts in other South Asian low-resource languages, such as Hindi and Tamil, have demonstrated the effectiveness of cross-lingual and transformer-based models for sentiment analysis [8], [9]. However, Bangla remains underrepresented in such studies despite its large speaker base. Our work addresses this gap by applying a multilingual transformer (XLM-RoBERTa) specifically to crime-related sentiment in Bangla, contributing both technically and contextually to the field.

The remainder of this paper is structured as follows: Section II reviews related work in sentiment analysis and transformer-based models. Section III details the proposed methodology, covering dataset construction, model architecture, and training procedures. Section IV presents experimental results and discussion. Finally, Section V concludes the paper, summarizing key findings and suggesting future research directions.

## II. Related Work

Early Bengali sentiment analysis studies ran upon a major constraint: a lack of labelled data. [1] undertook a first effort compiling 13,509 user comments from a Bangladeshi news source, personally annotated for sentiment. Analyzing several models including SVM, CNN, and LSTM, the research found LSTM best with an F1-score of 79.29%. This underlined how well LSTM could pick contextual nuances in Bengali text. In line with this, Wahid et al. [2] concentrated on social media conversations on cricket, therefore confirming the potency of LSTM. They discovered data sparsity problems even with a 10,000-entry dataset, which Word2Vec embeddings helped to somewhat offset.

Several machine learning techniques have been investigated by researchers in order to maximize sentiment classification. Reviewing Bengali e-commerce data, Akhtar et al. [10] compared five methods. Emphasizing the need of balanced datasets and oversampling methods, K-Nearest Neighbors' (KNN) surprisingly beat others with 96.25% accuracy. Jahan et al. [3] on another task identifying abusive language in Facebook comments—achieved 72.4% accuracy, underscoring the increasing need for toxicity moderation tools. With LSTM applied to Bengali tweets for negative sentiment recognition, Uddin et al. [11] obtained 86.3% accuracy. Their work showed how feature engineering and hyperparameter manipulation might offset little training data. In another study, Soumik et al. [12] assessed 10,000 Bengali app reviews, where SVM ranked highest (76.48% accuracy), therefore demonstrating the continued viability of traditional models. Analyzing 2,000 movie reviews, Rahman and Hossain [13] used multinomial Naïve Bayes with 88.5% accuracy above SVM and decision trees when coupled with unigram and POS-tagging information.

Deep learning and hybrid architectures have lately taken the stage in recent developments. Using a multinomial neural network with 86.67% accuracy, Khan et al. [14] binary sentiment categories happy/sad, classified Bengali social media messages. Furthermore, improving interpretability, Kibatia Chowdhury et al. [15] merged Bangla-BERT with lexicon-based techniques for financial sentiment analysis. Islam and Alam [16] presented BangDSA—a dataset of 203,000 Bengali microblog comments and offered skipBangla-BERT, combining Bangla-BERT with Skip-gram embeddings a big jump. With 95.71% (3-class) and 90.24% (15-class) accuracy, their CNN-BiLSTM model set fresh benchmarks.

One important topic that has surfaced is aspect-based sentiment analysis (ABSA). Ma et al. [7] improved aspect-specific sentiment recognition by means of augmented LSTM with commonsense knowledge using Sentic LSTM. Adapting the SemEval 2014 restaurant dataset for Bengali, Haque et al. [17] found SVM and KNN most effective. Sajjad and Jayarathna [18] suggested cross-lingual self-supervised learning to generate pseudo-labels using machine translation, hence mitigating data scarcity. For low resource languages, this method provided a scalable solution by increasing F1-scores by 15–25% without significant manual annotations.

LSTM, CNN, and hybrid models have shown excellent results, hence Bengali sentiment analysis has advanced dramatically. Still, there are ongoing difficulties with informal language, code-mixing, and dataset limits.

## III. Methodology

To classify Bengali sentiment, we employed the XLM-RoBERTa Base [19] model. Data collection, preprocessing, augmentation, training, evaluation, and explainability are all part of the workflow. The entire workflow is depicted in the Figure 1.

### A. Data Collection

In the contemporary digital age, social media platforms generate a vast volume of data readily accessible on the internet. However, the amount of written data in Bengali pales in comparison to other languages. As a low-research language, collecting Bangla crime news data was challenging. To overcome the challenges, we targeted the comment sections of Bengali crime articles from different sources, such as Facebook pages, YouTube news reports, etc. Finally, we managed to collect 28,528 comment data for the study. Table I consists of the sample data.

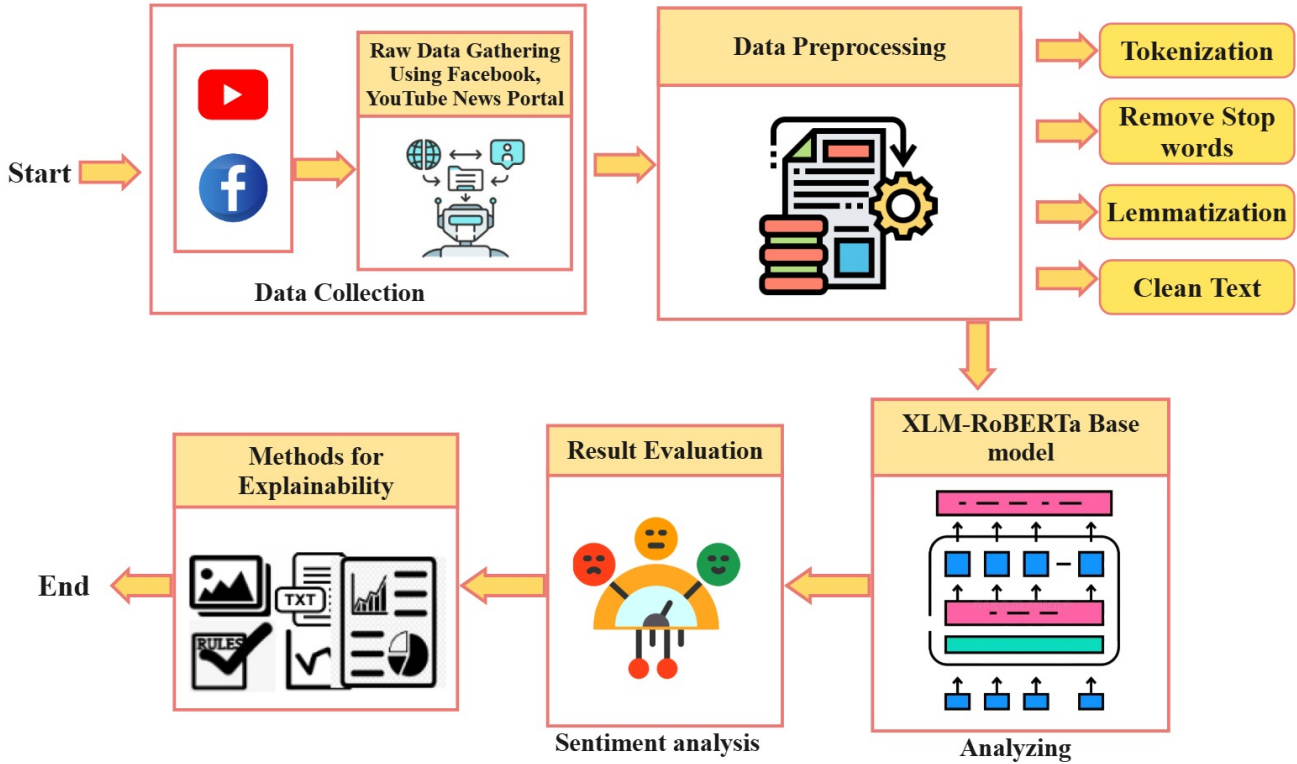

Figure 1: Overview of The Framework

Table I: Sample Examples of Comments on Recent Crime Events from The Dataset

| Comments | Source Information |
|---|---|
| এই ধোঁকাবাজি গুলো আজ থেকে নয় অনেক আগে থেকেই চলতেছে | Collected from Facebook page |
| কিন্তু যারা টাকা ইনভেস্ট করেছে তারা বেঁচে থেকেও মরে গেল। | Collected from YouTube news report |

Table II: Label Frequency in the Dataset

| Class | Count |
|---|---|
| Positive | 10,008 |
| Negative | 9,352 |
| Neutral | 9,168 |
| **Total** | **28,528** |

One limitation of the collected dataset is its potential lack of demographic and geographic diversity. Since the comments are sourced primarily from public social media platforms like Facebook and YouTube, they may over-represent urban users with internet access, leaving out opinions from rural or digitally disconnected populations.

### B. Dataset Description

The dataset is annotated for sentiment analysis, with each comment labeled as positive, neutral, and negative. Specifically, it contains 10,008 instances labeled as positive, 9,352 as negative, and the remaining entries correspond to the neutral category. With a good distribution across all classes, the dataset has 28528 data pieces in total. Table II. shows the Label Frequency in the Dataset.

### C. Data Preprocessing

After gathering the data, we checked the dataset to determine whether any empty rows were present We discard all the empty rows to address the missing data. Next, clean the data by removing special characters, numbers, and symbols to improve the results. After cleaning, the text was tokenized using the pretrained XLM-RoBERTa tokenizer, [19] which applies sub word segmentation via Byte-Pair Encoding (BPE) [20]. This approach preserves semantic integrity, handles rare words effectively, and supports multilingual input, which is particularly useful for processing Bengali text with code-mixed pattern.

To evaluate preprocessing efficacy, we conducted an ablation study comparing model performance on raw versus cleaned text. The cleaned data showed a 3.2% F1-score improvement, confirming our approach. Native Bengali speakers further verified 500 samples to ensure semantic integrity was preserved during cleaning and tokenization.

### D. Data Augmentation

Since there are 28528 observations in the dataset, enriching the data is one way to improve the outcome. Therefore, one approach to get around the restriction is to supplement the data. The data augmentation framework translates raw Bangla data to English after receiving it as input. The English data was then translated back to Bangla. We used the GoogleTrans package, which is free and available in Python, to enhance the data. Figure 2 illustrates an example.

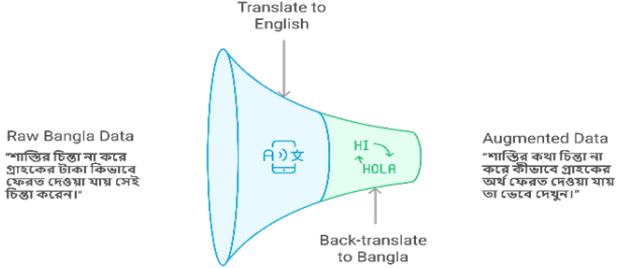

Figure 2: Data Augmentation

The AdamW optimization technique is used to train the suggested model, and it works well with transformer-based architectures. The major indicator used for performance evaluation is classification accuracy.

Table III: Parameter Details of The Proposed Model

| Hyperparameters | Values |
|---|---|
| Batch Size | 16 |
| Learning Rate | 0.00002 |
| Number of Epochs | 15 |
| Weight decay | 0.01 |
| Optimizer | AdamW |
| Loss Function | Cross-Entropy |
| Activation Function | GELU |

The hyperparameter values presented in Table III were determined through a constrained grid search conducted on a development set. To determine the best training setup, we ran a series of tests with various hyperparameter values. Three batch sizes—eight, sixteen, and thirty-two—were specifically examined. We examined learning rates at 1e-5, 2e-5, and 3e-5. The number of training epochs was also adjusted, with values of 10, 15, and 20. These combinations were thoroughly examined to discover which setup produces the highest validation performance. Given computational limitations, we prioritized both training efficiency and model generalization, ultimately selecting the hyperparameters that achieved the highest validation accuracy.

## IV. Experimental Results and Discussion

This part contains the data, as well as some comparative analysis and explanations for our suggested model. In this study, Python is used as the programming language to implement and Jupyter Notebook to execute all of our modules. Moreover, several libraries and packages have been used to perform different tasks, such as Pandas [15] for extracting information from the dataset, Matplotlib [6] for visualizing and plotting all the processed data, Numpy [5] for performing several mathematical functions, Scikit-learn [4] for splitting the dataset, generating classification reports and confusion matrix, and Keras [16] for building the models using the TensorFlow [18] library as the backend.

### A. Performance Evaluation

Four models were evaluated in this work: LSTM, LSTM boosted with an attention mechanism, BanglaBERT, and a custom-designed recommended model in order to test the usefulness of numerous architectures in Bangla sentiment classification. Every model was trained and validated and assessed depending on training, validation, and testing accuracy using the same dataset.

Table IV: Accuracy of Different Models

| Model | Train | Validation | Test |
|---|---|---|---|
| LSTM | 97.0% | 93.4% | 93.4% |
| LSTM + Attention | 98.3% | 94.2% | 94.2% |
| BanglaBERT | 92.6% | 91.1% | 91.2% |
| Proposed Model | 98.6% | 97.0% | 97.0% |

Table V: Sentiment Classification Scores

| Class | Precision | Recall | F1-score | Support |
|---|---|---|---|---|
| Neutral | 0.98 | 0.96 | 0.97 | 917 |
| Positive | 0.95 | 0.98 | 0.96 | 1001 |
| Negative | 0.96 | 0.95 | 0.95 | 935 |
| Accuracy | | | 0.97 | |
| Macro Avg | 0.97 | 0.96 | 0.97 | 2853 |
| Weighted Avg | 0.97 | 0.97 | 0.97 | 2853 |

This study evaluated four architectures for Bangla sentiment classification: LSTM, LSTM with Attention, BanglaBERT, and a proposed custom model. The baseline LSTM achieved 97% training accuracy and 93.4% on both validation and test sets, indicating slight overfitting. Incorporating an attention mechanism improved the model's performance, resulting in 98.3% training accuracy and 94.2% accuracy on both validation and test sets, which indicates better generalization. BanglaBERT, despite being a transformer-based model pre-trained for Bangla, recorded slightly lower performance with 92.6% training, 91.1% validation, and 91.2% test accuracy likely due to domain mismatch or fine-tuning limitations.

The proposed model outperformed all others, achieving 98.56% training and 97% on both validation and testing. The minimal performance gap indicates a strong generalization, likely due to architectural optimizations such as enhanced embeddings, attention layers, or regularization techniques. In Table V, the F1 score for all three classes is above 95%. Nonetheless, the model's total accuracy is 97%.

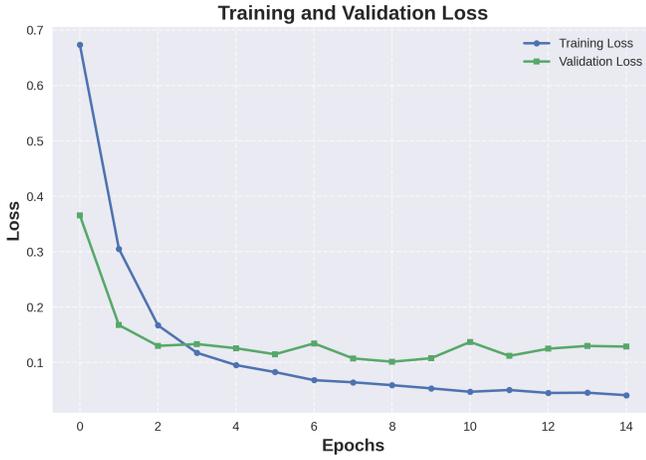
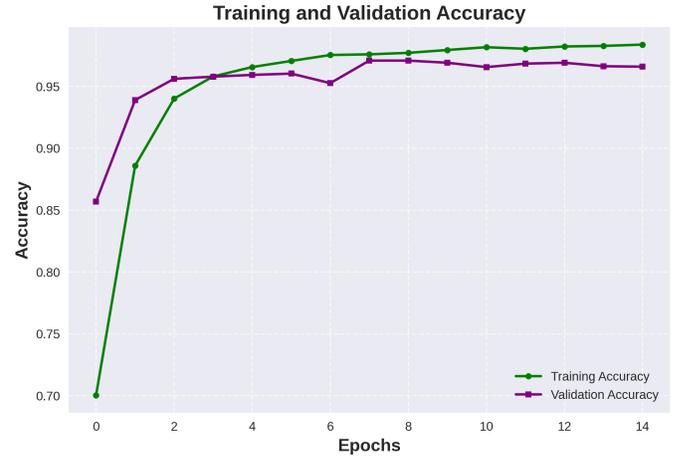

Figure 3: Graph of accuracy and loss for the proposed model

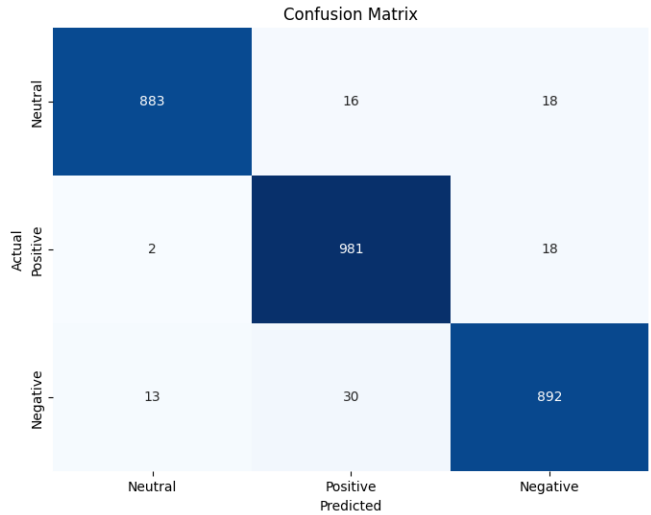

Figure 4: Confusion Matrix

Figure 4's confusion matrix reveals that the model confuses positive and negative data, as compared to others.

এমন হত্যাকান্ড মেনে নিতে কষ্ট হয়।

This statement is a negative one, however model predict it as positive. In such cases, the model gets confused and provides lower accuracy.

### B. Explainability Analysis

This section aims to examine the underlying factors contributing to the model's predictions as part of the explainability analysis. Explainability techniques of deep learning, often referred to as XAI techniques [7], allow us to give insight into how the models generate results. We can analyze the output of the model using XAI techniques, which consequently builds conviction in those results.

An explainer object was built using the `explainer.explain_instance` function to discover the important features impacting individual projected outcomes. This strategy improves interpretability by highlighting the most important phrases or attributes that contribute to the model's outputs.

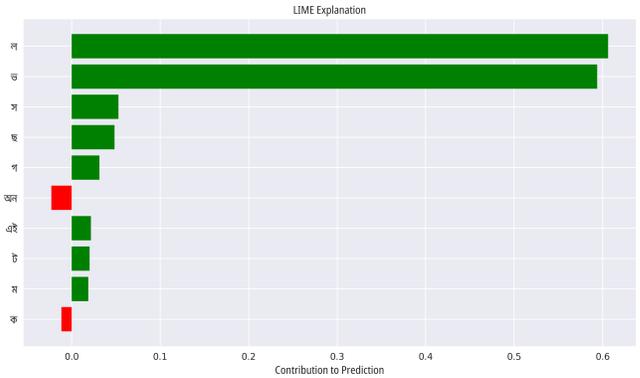

Figure 5: LIME Output

Table VI: Comparison with Advanced Models in Bengali Sentiment Analysis

| **Study** | **Model** | **Accuracy** |
|---|---|---|
| Akhtar et al. [10] | KNN | 96.25% |
| Pran et al. [19] | LSTM | 95.33% |
| Islam & Alam [21] | CNN-BiLSTM | 95.71% |
| **Proposed** | **XLM-RoBERTa** | **97.00%** |

In the Figure 5, the specific character prediction probability is way higher than others. The LIME also shows the influential character " ল, ভ, স, ছ, গ, এই, ট, ম " for the prediction.

The proposed model achieves 97% accuracy, outperforming prior methods like KNN, LSTM, and so on. It shows in the Table VI, and Figure 3 illustrates the graph of accuracy and loss respectively.

## V. Conclusion and Future Work

In this work, the authors propose a transformer-based model, XLM-RoBERTa Base, for identifying public sentiment expressed in comments related to crime news. The study will concentrate on the transformer-based model, which has undergone validation using a dataset of 28528 comments from various online platforms. After training the model, it works satisfactorily for all the sentiment classes, with an average F1 score of 97%.

The suggested model is carefully compared against various modern deep learning algorithms in order to determine its relative performance. In addition, Explainable AI (XAI) approaches are used to investigate the model's sentiment predictions, which improves interpretability. In the near future, the authors aim to analyse sentiment using multimodal dataset.

Future work should also consider expanding the dataset to include a more geographically and demographically representative sample of the population, potentially through collaboration with local agencies or surveys in different regions.

## References

[1] M. A.-U.-Z. Ashik, S. Shovon, and S. Haque, "Data set for sentiment analysis on bengali news comments and its baseline evaluation," in *2019 International conference on bangla speech and language processing (ICBSLP)*. IEEE, 2019, pp. 1–5.

[2] M. F. Wahid, M. J. Hasan, and M. S. Alom, "Cricket sentiment analysis from bangla text using recurrent neural network with long short term memory model," in *2019 International Conference on Bangla Speech and Language Processing (ICBSLP)*. IEEE, 2019, pp. 1–4.

[3] M. Jahan, I. Ahamed, M. R. Bishwas, and S. Shatabda, "Abusive comments detection in bangla-english code-mixed and transliterated text," in *2019 2nd international conference on innovation in engineering and technology (ICIET)*. IEEE, 2019, pp. 1–6.

[4] F. Pedregosa, G. Varoquaux, A. Gramfort, V. Michel, B. Thirion, O. Grisel, M. Blondel, P. Prettenhofer, R. Weiss, V. Dubourg *et al.*, "Scikit-learn: Machine learning in python," *the Journal of machine Learning research*, vol. 12, pp. 2825–2830, 2011.

[5] C. R. Harris, K. J. Millman, S. J. Van Der Walt, R. Gommers, P. Virtanen, D. Cournapeau, E. Wieser, J. Taylor, S. Berg, N. J. Smith *et al.*, "Array programming with numpy," *Nature*, vol. 585, no. 7825, pp. 357–362, 2020.

[6] J. D. Hunter, "Matplotlib: A 2d graphics environment," *Computing in science & engineering*, vol. 9, no. 03, pp. 90–95, 2007.

[7] S. M. Lundberg and S.-I. Lee, "A unified approach to interpreting model predictions," *Advances in neural information processing systems*, vol. 30, 2017.

[8] S. Sazzed, "Improving sentiment classification in low-resource bengali language utilizing cross-lingual self-supervised learning," in *International conference on applications of natural language to information systems*. Springer, 2021, pp. 218–230.

[9] P. Joshi, S. Santy, A. Budhiraja, K. Bali, and M. Choudhury, "The state and fate of linguistic diversity and inclusion in the nlp world," *arXiv preprint arXiv:2004.09095*, 2020.

[10] M. T. Akter, M. Begum, and R. Mustafa, "Bengali sentiment analysis of e-commerce product reviews using k-nearest neighbors," in *2021 International conference on information and communication technology for sustainable development (ICICT4SD)*. IEEE, 2021, pp. 40–44.

[11] A. H. Uddin, D. Bapery, and A. S. M. Arif, "Depression analysis from social media data in bangla language using long short term memory (lstm) recurrent neural network technique," in *2019 international conference on computer, communication, chemical, materials and electronic engineering (IC4ME2)*. IEEE, 2019, pp. 1–4.

[12] M. M. J. Soumik, S. S. M. Farhavi, F. Eva, T. Sinha, and M. S. Alam, "Employing machine learning techniques on sentiment analysis of google play store bangla reviews," in *2019 22nd International Conference on Computer and Information Technology (ICCIT)*. IEEE, 2019, pp. 1–5.

[13] A. Rahman and M. S. Hossen, "Sentiment analysis on movie review data using machine learning approach," in *2019 international conference on bangla speech and language processing (ICBSLP)*. IEEE, 2019, pp. 1–4.

[14] G. Bradski, "The opencv library." *Dr. Dobb's Journal: Software Tools for the Professional Programmer*, vol. 25, no. 11, pp. 120–123, 2000.

[15] W. McKinney *et al.*, "Data structures for statistical computing in python." *SciPy*, vol. 445, no. 1, pp. 51–56, 2010.

[16] F. Chollet *et al.*, "Keras: Deep learning library for theano and tensorflow," *URL: https://keras. io/k*, vol. 7, no. 8, p. T1, 2015.

[17] D. Kang and Y. Park, "based measurement of customer satisfaction in mobile service: Sentiment analysis and vikor approach," *Expert Systems with Applications*, vol. 41, no. 4, pp. 1041–1050, 2014.

[18] M. Abadi, A. Agarwal, P. Barham, E. Brevdo, Z. Chen, C. Citro, G. S. Corrado, A. Davis, J. Dean, M. Devin *et al.*, "Tensorflow: Large-scale machine learning on heterogeneous systems (2015), software available from tensorflow. org," 2019.

[19] M. S. A. Pran, M. R. Bhuiyan, S. A. Hossain, and S. Abujar, "Analysis of bangladeshi people's emotion during covid-19 in social media using deep learning," in *2020 11th International Conference on Computing, Communication and Networking Technologies (ICCCNT)*. IEEE, 2020, pp. 1–6.

[20] X. Han, J. Wang, M. Zhang, and X. Wang, "Using social media to mine and analyze public opinion related to covid-19 in china," *International journal of environmental research and public health*, vol. 17, no. 8, p. 2788, 2020.

[21] M. S. Islam and K. M. Alam, "Sentiment analysis of bangla language using a new comprehensive dataset bangdsa and the novel feature metric skipbangla-bert," *Natural Language Processing Journal*, vol. 7, p. 100069, 2024.